\documentclass[runningheads]{llncs}

\def \cD{{\mathcal{D}}}

\def \cH{{\mathcal{H}}}

\def \bbR{{\mathbb{R}}}

\def \bA{{\mathbf{A}}}
\def \bD{{\mathbf{D}}}

\def \bI{{\mathbf{I}}}
\def \bK{{\mathbf{K}}}
\def \bM{{\mathbf{M}}}
\def \bX{{\mathbf{X}}}
\def \bZ{{\mathbf{Z}}}
\def \bL{{\mathbf{L}}}

\def \bU{{\mathbf{U}}}

\newcommand{\bw}{{\mathbf{w}}}
\newcommand{\bW}{{\mathbf{W}}}
\newcommand{\bx}{{\mathbf{x}}}
\newcommand{\by}{{\mathbf{y}}}
\newcommand{\bz}{{\mathbf{z}}}

\def \tr{{\mathrm{tr}}}

\usepackage{float}
\usepackage{algorithm}
\usepackage[noend]{algpseudocode}
\usepackage{colortbl}
\usepackage{tablefootnote}
\usepackage{amsfonts}
\usepackage{mathrsfs}
\usepackage{amsmath}
\usepackage{mathtools}
\usepackage{tabularx}

\usepackage{caption}
\usepackage{subfig}
\usepackage{multirow}
\usepackage{hhline}
\usepackage{multicol}

\usepackage{graphicx}
\usepackage{amsmath,amssymb} % define this before the line numbering.
\usepackage{color}
\usepackage[width=122mm,left=12mm,paperwidth=146mm,height=193mm,top=12mm,paperheight=217mm]{geometry}
\begin{document}
% \linenumbers
\pagestyle{headings}
\mainmatter

\title{Nonlinear Embedding Transform for Unsupervised Domain Adaptation} % Replace with your title
\author{Hemanth Venkateswara, Shayok Chakraborty, Sethuraman Panchanathan}
%Please write out author names in full in the paper, i.e. full given and family names. 
%If any authors have names that can be parsed into FirstName LastName in multiple ways, please include the correct parsing, in a comment to the volume editors:
%\index{Lastnames, Firstnames}
%(Do not uncomment it, because you may introduce extra index items if you do that...)
\institute{Center for Cognitive Ubiquitous Computing\\
	Arizona State University, AZ, USA\\
	\email{ \{hemanthv,schakr10,panch\}@asu.edu}
}

\maketitle

\begin{abstract}
The problem of domain adaptation (DA) deals with adapting classifier models trained on one data distribution to different data distributions. 
In this paper, we introduce the Nonlinear Embedding Transform (NET) for unsupervised DA by combining domain alignment along with similarity-based embedding. 
We also introduce a validation procedure to estimate the model parameters for the NET algorithm using the source data. 
Comprehensive evaluations on multiple vision datasets demonstrate that the NET algorithm outperforms existing competitive procedures for unsupervised DA.
\end{abstract}

\section{Introduction}
%Most vision based multimedia applications that require some form of supervised training depend upon labeled data, which is expensive to obtain. 
% Labeled data is expensive to obtain but there are large volumes of unlabeled data available online owing to the exponential increase in the number of images and videos uploaded online. 
Classification models trained on labeled datasets are ineffective over data from different distributions owing to data-shift \cite{torralba2011unbiased}. 
The problem of domain adaptation (DA) deals with adapting models trained on one data distribution (source domain) to different data distributions (target domains). 
%Owing to the exponential increase in the amount of unlabeled data available online, there is a compelling demand to build domain-adaptive models to analyze multimedia content. 
%Supervised DA techniques assume the presence of a few labels for the target data \cite{aytar2011tabula}, \cite{duan2012domain}, \cite{hoffman2013}, \cite{venkateswara2015coupled}, when training a DA classifier. 
%Real world applications need not support labeled data in the target domain and adaptation here is termed as unsupervised DA. 
For the purpose of this paper, we organize unsupervised DA procedures under two categories: \textit{linear} and \textit{nonlinear}, based on feature representations used in the model. 
\textit{Linear} techniques determine linear transformations of the source (target) data and align it with the target (source), or learn a linear classifier with the source data and adapt it to the target data \cite{bruzzone2010domain}, \cite{sun2015return} and \cite{venkateswara2015coupled}. 
%\textit{Nonlinear} techniques are deployed in certain applications where the source and target domains cannot be aligned using linear transformations.  
\textit{Nonlinear} procedures on the other hand, apply nonlinear transformations to reduce cross-domain disparity \cite{gong2012geodesic}, \cite{pan2011domain}. 

In this work we present the Nonlinear Embedding Transform (NET) procedure for unsupervised DA. 
The NET consists of two steps, (i) Nonlinear domain alignment using Maximum Mean Discrepancy (MMD) \cite{long2013transfer}, (ii) similarity-based embedding to cluster the data for enhanced classification. 
In addition, we introduce a procedure to sample source data in order to generate a validation set for model selection. 
We study the performance of the NET algorithm with popular DA datasets for computer vision. 
Our results showcase significant improvement in the classification accuracies compared to competitive DA procedures. 

\section{Related Work}
In this section we provide a concise review of some of the unsupervised DA procedures closely related to the NET. 
Under \textit{linear} methods, Burzzone et al. \cite{bruzzone2010domain}, proposed the DASVM algorithm to iteratively adapt a SVM trained on the source data, to the unlabeled target data. 
The state-of-the-art linear DA procedures are Subspace Alignment (SA), by Fernando et al. \cite{fernando2013unsupervised}, and the CORAL algorithm, by Sun et al. \cite{sun2015return}. 
The SA aligns the subspaces of the source and the target with a linear transformation and the CORAL transforms the source data such that the covariance matrices of the source and target are aligned.  

\textit{Nonlinear} procedures generally project the data to a high-dimensional space and align the source and target distributions in that space.  
The popular GFK algorithm by Gong et al. \cite{gong2012geodesic}, projects the two distributions onto a manifold and learns a transformation to align them. 
%Wang et al. \cite{wang2011heterogeneous}, and Tuia et al. \cite{tuia2016kernel}, proposed manifold alignment procedures closely related to the NET. However, these are supervised procedures that break down when the target has no labels. 
The Transfer Component Analysis (TCA) \cite{pan2011domain}, Transfer Joint Matching (TJM) \cite{long2014transfer}, and Joint Distribution Adaptation (JDA) \cite{long2013transfer}, algorithms, apply MMD-based projection to nonlinearly align the domains. 
In addition, the TJM implements instance selection using $\ell_{2,1}$-norm regularization and the JDA performs a joint distribution alignment of the source and target domains. 
The NET implements nonlinear alignment of the domains along with a similarity preserving projection, which ensures that the projected data is clustered based on category. 
We compare the NET with only kernel-based nonlinear methods and do not include deep learning based DA procedures.  

\section{DA With Nonlinear Embedding}
In this section we outline the problem of unsupervised DA and develop the NET algorithm. 
%We consider two domains; source domain $\sS$ and target domain $\sT$. Let $\cD_s = \{(\bx_i^s, y_i^s)\}_{i=1}^{n_s} \subset \sS$ be a subset of the source domain and $\cD_t = \{(\bx_i^t, y_i^t)\}_{i=1}^{n_t} \subset \sT$ be the subset of the target domain. 
Let $\bX_S = [\bx_1^s, \ldots, \bx_{n_s}^s] \in \bbR^{d\times n_s}$ and $\bX_T = [\bx_1^t, \ldots, \bx_{n_t}^t] \in \bbR^{d\times n_t}$ be the source and target data points respectively. 
Let $Y_S = [y_1^s, \ldots, y_{n_s}^s]$ and $Y_T = [y_1^t, \ldots, y_{n_t}^t]$ be the source and target labels respectively. 
Here, $\bx_i^s$ and $\bx_i^t$ $\in \bbR^d$ are data points and $y_i^s$ and $y_i^t$ $\in \{1,\ldots,C\}$ are the associated labels. 
We define $\bX \coloneqq [\bx_1, \ldots, \bx_n] = [\bX_S, \bX_T]$, where $n = n_s + n_t$. 
In the case of unsupervised DA, the labels $Y_T$ are missing and the joint distributions for the two domains are different, i.e. $P_S(X,Y) \neq P_T(X,Y)$. 
The task lies in learning a classifier $f(\bx) = p(y|\bx)$, that predicts the labels of the target data points. 
%The problem of DA deals with the situation where the joint distributions for the source and target domains are different, i.e. $P_S(X,Y) \neq P_T(X,Y)$. 
%In the case of unsupervised DA, the labels $Y_T$ are missing and the task lies in learning a classifier $f(\bx) = p(y|\bx)$ to predict the labels of the target data points, where $p(y|\bx)$ is the posterior probability of label $y$ given data point $\bx$. 
%The classifier is applied to estimate the labels of the target data $\hat{Y}_T = [\hat{y}_1^t, \ldots, \hat{y}_{n_t}^t]$ corresponding to $\bX_T$ using $\cD_s$ and $\bX_T$. 
%Since the joint distributions of the source and target domains are different, a classifier trained using only the source data $\cD_s$ may not accurately predict the labels for the target data. 
%The NET algorithm for unsupervised DA modifies $\bX_S$ and $\bX_T$ using a nonlinear transformation and embeds the data points in a subspace that clusters data points having the same labels. The transformed and target-aligned source data is used to train a classifier, which can predict the target labels. 

%\textbf{Notation}: $\bI$ refers to the identity matrix of appropriate size. $\mathbf{1}$ refers to a matrix of ones. Vectors are represented by bold lowercase letters (for example $\bu$, $\bx$). Bold upper case letters denote matrices (for example $\bA$, $\bK$). 
%$||\bA||_F$  refers to the Frobenius  norm of matrix $\bA$. $\tr(\bA)$ and $\bA^\top$ refer to the trace and transpose of $\bA$ respectively. 

\subsection{Nonlinear Embedding for DA}
One of the techniques to reduce domain disparity is to project the source and target data to a common subspace. 
KPCA is a popular nonlinear projection algorithm where data is first mapped to a high-dimensional (possibly infinite-dimensional) space given by $\Phi(\bX) = [\phi(\bx_1), \ldots, \phi(\bx_n)]$. 
$\phi:\bbR^d \rightarrow \cH$ defines the mapping and $\cH$ is a RKHS with a psd kernel $k(\bx,\by) = \phi(\bx)^\top\phi(\by)$. 
The kernel matrix for $\bX$ is given by $\bK = \Phi(\bX)^\top\Phi(\bX) \in \bbR^{n\times n}$. 
The mapped data is then projected onto a subspace of eigen-vectors (directions of maximum nonlinear variance in the RKHS). 
The top $k$ eigen-vectors in the RKHS are obtained using the representor theorem, $\bU = \Phi(\bX)\bA$, where $\bA \in \bbR^{n\times k}$ is the matrix of coefficients that needs to be determined. 
The nonlinearly projected data is then given by, $\bZ = [\bz_1, \ldots, \bz_n] = \bA^\top\bK \in \bbR^{k\times n}$, where $\bz_i \in \bbR^k$, $i = 1,\ldots,n$, are the projected data points. 

In order to reduce the domain discrepancy in the projected space, we implement the joint distribution adaptation (JDA), as outlined in \cite{long2013transfer}. 
The JDA seeks to align the marginal and conditional probability distributions of the projected data ($\bZ$), by estimating the coefficient matrix $\bA$, which minimizes: 
\begin{footnotesize}
	\begin{flalign}
		\min_{\bA} \sum_{c=0}^C\tr(\bA^\top\bK\bM_c\bK^\top\bA).
		\label{Eq:JDA}
	\end{flalign}
\end{footnotesize}
\noindent $\tr(.)$ refers to trace and $\bM_c$, where $c = 0,\ldots,C$, are $n \times n$ matrices given by, 
\begin{footnotesize}
\raggedleft
\begin{minipage}{.49\textwidth}
  \begin{flalign}
    (M_c)_{ij} = 
		\begin{cases}
				\frac{1}{n_s^{(c)}n_s^{(c)}},& \bx_i, \bx_j \in \cD_s^{(c)}\\
				\frac{1}{n_t^{(c)}n_t^{(c)}},& \bx_i, \bx_j \in \cD_t^{(c)}\\
				\frac{-1}{n_s^{(c)}n_t^{(c)}},& \begin{cases} 
																						\bx_i \in \cD_s^{(c)}, \bx_j \in \cD_t^{(c)}\\
																						\bx_j \in \cD_s^{(c)}, \bx_i \in \cD_t^{(c)}\\
																				\end{cases}\\
				0, & \text{otherwise},
		\end{cases}
  \end{flalign}
	\end{minipage}
	\raggedright
	\begin{minipage}{.49\textwidth}
	\begin{flalign}
    (M_0)_{ij} = 
		\begin{cases}
				\frac{1}{n_sn_s},& \bx_i, \bx_j \in \cD_s\\
				\frac{1}{n_tn_t},& \bx_i, \bx_j \in \cD_t\\
				\frac{-1}{n_sn_t},& \text{otherwise},\\
		\end{cases}
  \end{flalign}
\end{minipage}
\end{footnotesize}

\noindent where $\cD_s$ and $\cD_t$ are the sets of source and target data points respectively. 
$\cD_s^{(c)}$ is the set of source data points belonging to class $c$ and $n_s^{(c)} = |\cD_s^{(c)}|$. 
Likewise, $\cD_t^{(c)}$ is the set of target data points belonging to class $c$ and $n_t^{(c)} = |\cD_t^{(c)}|$. 
Since the target labels are unknown, we use predicted labels for the target data points. 
We begin with predicting the target labels using a classifier trained on the source data and refine these labels over iterations, to arrive at the final prediction. 
For more details please refer to \cite{long2013transfer}. 

In addition to domain alignment, we would like the projected data $\bZ$, to be classification friendly (easily classifiable). 
To this end, we introduce Laplacian eigenmaps to ensure a similarity-preserving projection such that data points with the same class label are clustered together. 
The similarity relations are captured by the $(n\times n)$ adjacency matrix $\bW$, and the optimization problem estimates the projected data $\bZ$; 

\begin{footnotesize}
\raggedleft
 \begin{minipage}{.49\textwidth}
  \begin{flalign}
    \bW_{ij} \coloneqq
			\begin{cases}
				1 &~~~ y_i^s=y_j^s ~ \text{or} ~ i=j\\
				0 & y_i^s\neq y_j^s ~\text{or labels unknown,}
			\end{cases}
			\label{Eq:Adj}
  \end{flalign}
 \end{minipage}
\raggedright
 \begin{minipage}{.49\textwidth}
  \begin{flalign}
    \min_{\bZ} \frac{1}{2}\sum_{ij}\bigg|\bigg|\frac{\bz_i}{\sqrt{d_i}} - \frac{\bz_j}{\sqrt{d_j}}\bigg|\bigg|^2\bW_{ij} = \tr(\bZ\bL\bZ^\top).
		\label{Eq:Lapl}
  \end{flalign}
 \end{minipage}
\end{footnotesize}

\noindent$\bD = \textit{diag}(d_1, \ldots, d_n)$ is the $(n\times n)$ diagonal matrix where, $d_i = \sum_j\bW_{ij}$ and $\bL$ is the normalized graph laplacian matrix that is symmetric positive semidefinite and is given by $\bL = \bI-\bD^{-1/2}\bW\bD^{-1/2}$, where $\bI$ is an identity matrix. 
When $\bW_{ij} = 1$, the projected data points $\bz_i$ and $\bz_j$ are close together (as they belong to the same category). 
The normalized distance between the vectors $||\bz_i/\sqrt{d_i} - \bz_j/\sqrt{d_j}||^2$, captures a more robust measure of data point clustering compared to the un-normalized distance $||\bz_i - \bz_j||^2$, \cite{chung1997spectral}.
\vspace{-4mm}
\subsection{Optimization Problem}
\vspace{-1mm}
\label{Sec:OptProb}
The optimization problem for NET is obtained from (\ref{Eq:JDA}) and (\ref{Eq:Lapl}) by substituting, $\bZ = \bA^\top\bK$. 
Along with regularization and a constraint, we get, 
\begin{footnotesize}
	\begin{flalign}
		\min_{\bA^\top\bK\bD\bK^\top\bA = \bI} \alpha.\tr(\bA^\top\bK\sum_{c=0}^C\bM_c\bK^\top\bA) + \beta.\tr(\bA^\top\bK \bL\bK^\top\bA) + \gamma||\bA||_F^2.
		\label{Eq:Opt}
	\end{flalign}
\end{footnotesize}
\noindent$\bA \in \bbR^{n\times k}$ is the projection matrix. 
The regularization term $||\bA||_F^2$ (Frobenius norm), controls the smoothness of projection and the magnitudes of $(\alpha, \beta, \gamma)$, denote the importance of the individual terms in (\ref{Eq:Opt}). 
The constraint prevents the data points from collapsing onto a subspace of dimensionality less than $k$, \cite{belkin2003laplacian}. 
%The Lagrangian for (\ref{Eq:Opt}) is given by,
%\begin{footnotesize}
	%\begin{flalign}
		%L(\bA, \mathbf{\Lambda)} = & \alpha.\tr\big(\bA^\top\bK \sum_{c=0}^C\bM_c\bK^\top\bA\big) + \beta.\tr(\bA^\top\bK \bL\bK^\top\bA) \notag  \\
															 %&+ \gamma||\bA||_F^2 + \tr((\bI - \bA^\top\bK\bD\bK^\top\bA)\mathbf{\Lambda}),
		%\label{Eq:Lagran}
	%\end{flalign}
%\end{footnotesize}
%\noindent where, $\mathbf{\Lambda} = \textit{diag}(\lambda_1, \ldots, \lambda_k)$, is the diagonal matrix of Lagrangian constants. 
%Setting $\frac{\partial L}{\partial \bA} = 0$, we obtain,
Equation (\ref{Eq:Opt}) can be solved by constructing the Lagrangian $L(\bA, \mathbf{\Lambda)}$, where, $\mathbf{\Lambda} = \textit{diag}(\lambda_1, \ldots, \lambda_k)$, is the diagonal matrix of Lagrangian constants (see \cite{long2014transfer}). 
Setting the derivative $\frac{\partial L}{\partial \bA} = 0$, yields the generalized eigen-value problem, 
\begin{footnotesize}
	\begin{flalign}
		\big(\alpha\bK \sum_{c=0}^C\bM_c \bK^\top + \beta\bK \bL\bK^\top + \gamma\bI\big)\bA = \bK\bD\bK^\top\bA\mathbf{\Lambda}.
		\label{Eq:GenEigen}
	\end{flalign}
\end{footnotesize}
\noindent $\bA$ is the matrix of the $k$-smallest eigen-vectors of (\ref{Eq:GenEigen}) and $\mathbf{\Lambda}$ is the diagonal matrix of eigen-values. 
The projected data points are given by, $\bZ = \bA^\top\bK$. 

\subsection{Model Selection}
\label{Sec:ModSel}
Current DA methods use the target data to validate the optimum parameters for their models \cite{long2014transfer}, \cite{long2013transfer}. 
%When target labels are not available, as in the case of real world DA applications, there is no way to verify if the parameters in these DA models are optimized. 
We introduce a new technique to evaluate $(\alpha, \beta, \gamma, k)$, using a subset of the source data as a validation set. 
The subset is selected by weighting the source data points using Kernel Mean Matching (KMM). 
%This technique has been used to weight source data points to reduce the distribution difference between source and target data \cite{fernando2013unsupervised}, \cite{gong2013connecting}.
The KMM computes source instance weights $w_i$, by minimizing, $\big|\big|\frac{1}{n_s}\sum_{i=1}^{n_s}w_i\phi(\bx_i^s) -  \frac{1}{n_t}\sum_{j=1}^{n_t}\phi(\bx_i^t)\big|\big|_\mathcal{H}^2$. 
Defining $\kappa_i := \frac{n_s}{n_t}\sum_{j=1}^{n_t}k(\bx_i^s, \bx_j^t)$, $i = 1,\ldots,n_s$ and $\bK_{S_{ij}} = k(\bx_i^s, \bx_j^s)$, the minimization can be written in terms of quadratic programming: 
\begin{footnotesize}
	\begin{flalign}
		\min_{\bw} = \frac{1}{2} \bw^\top\bK_S\bw - \mathbf{\kappa}^\top\bw, ~~\text{s.t.}~ w_i\in[0, B], ~\bigg|\sum_{i=1}^{n_s}w_i - n_s \bigg|\leq n_s\epsilon.
		\label{Eq:KMM}
	\end{flalign}
\end{footnotesize}
The first constraint limits the scope of discrepancy between source and target distributions with $B\rightarrow 1$, leading to an unweighted solution. 
The second constraint ensures the measure $w(x)P_S(x)$, is a probability distribution \cite{gretton2009covariate}. 
In our experiments, the validation set is 30\% of the source data with the largest weights. 
This validation set is used to estimate the best values for $(\alpha, \beta, \gamma, k)$. 
%For fixed values of $(\alpha, \beta, \gamma, k)$, the NET model is trained using the source data (without the validation set) and target data. 
%The model is tested on the validation set to estimate parameters yielding highest classification accuracies. 
\begin{figure}[t]
\centering
\subfloat[\scriptsize{\# bases $k$}]{
		\label{Fig:kStudy}%[0.23\textwidth] % width of caption
    \includegraphics[trim = 2mm 0mm 11mm 0mm, clip, width=0.23\textwidth]{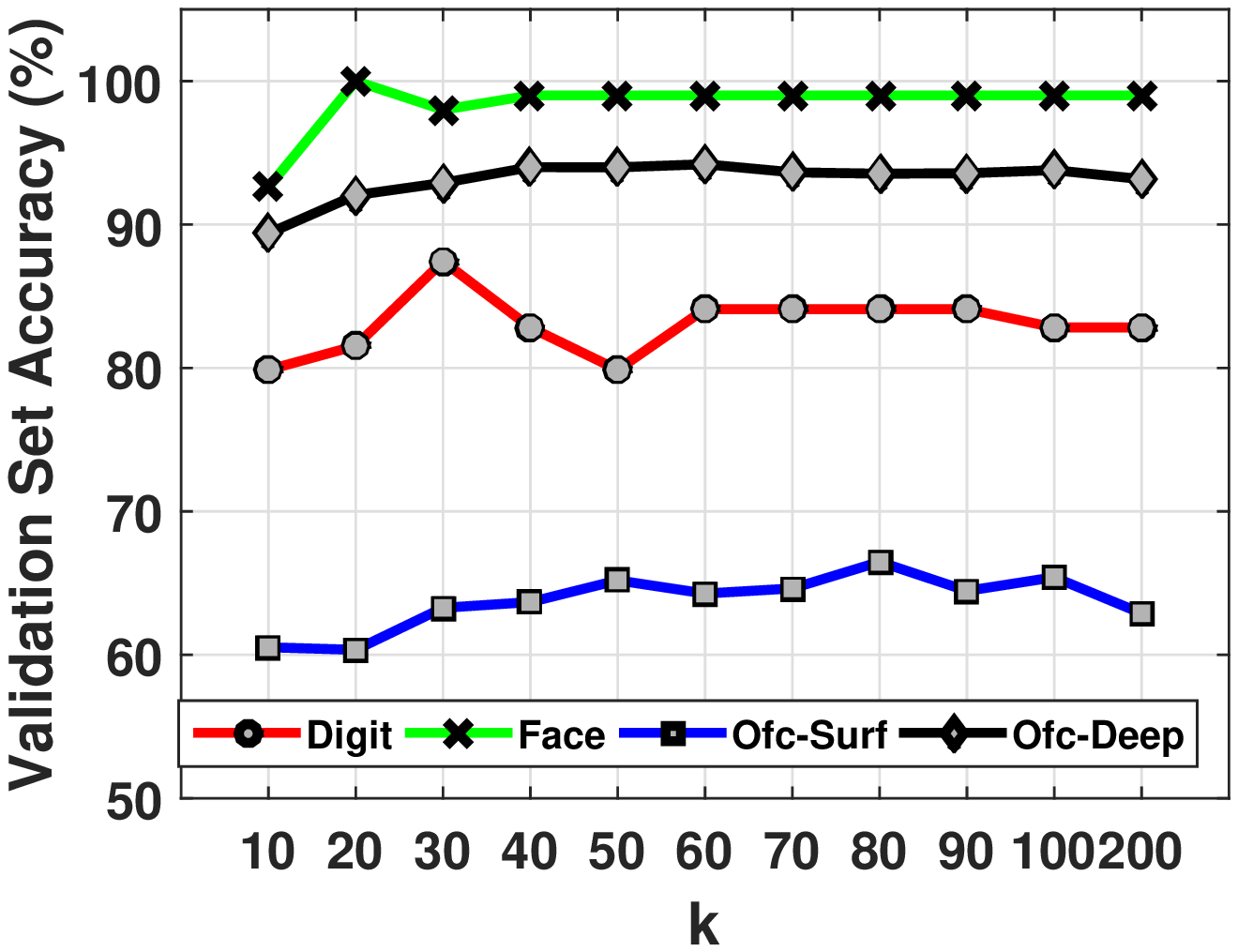}
}%
\hfill
%\hspace{0.01\textwidth} % seperation
\subfloat[\scriptsize{MMD weight $\alpha$}]{
		\label{Fig:aStudy}%[0.23\textwidth] % width of caption
    \includegraphics[trim = 2mm 0mm 11mm 0mm, clip, width=0.23\textwidth]{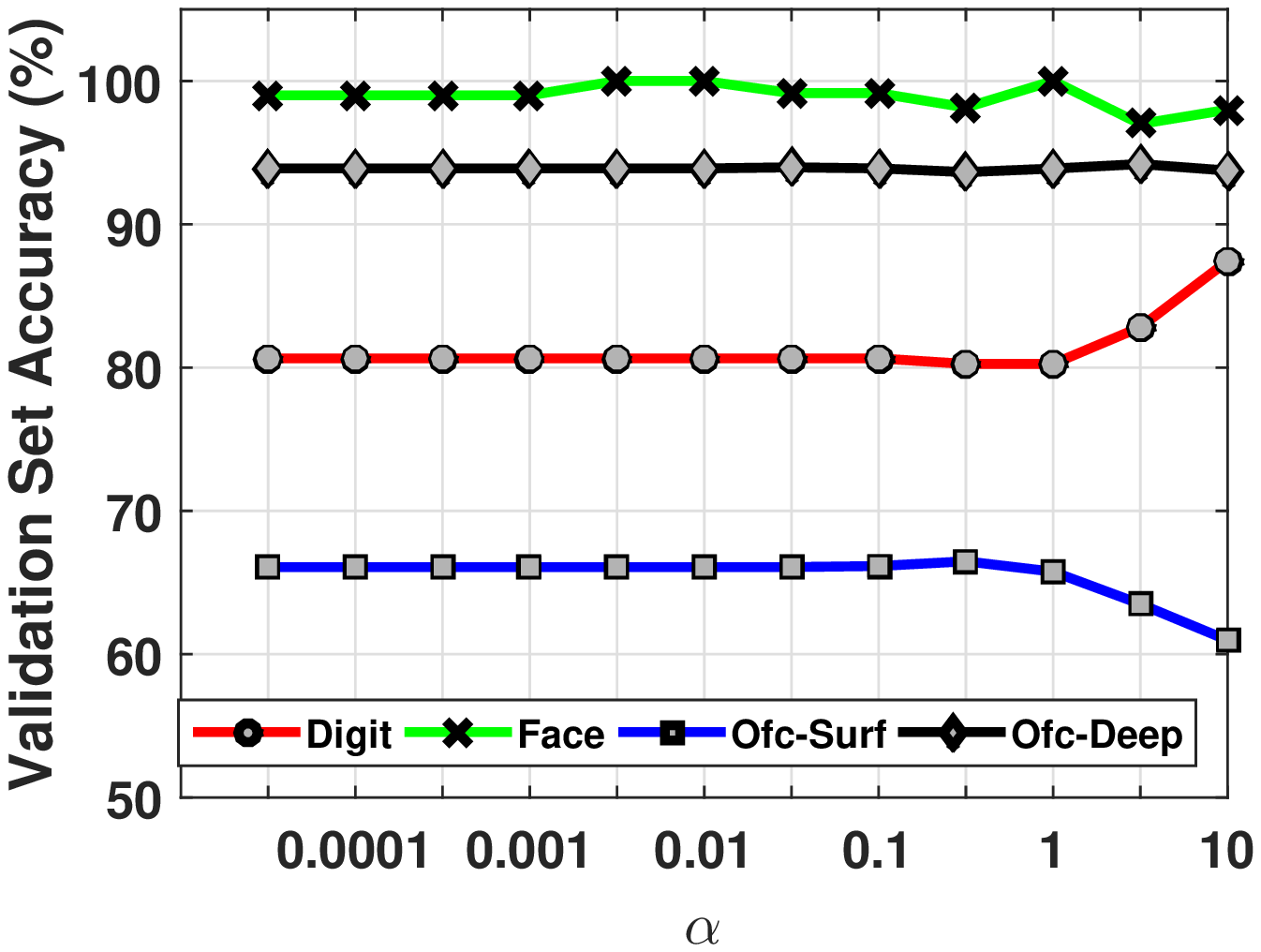}
}%
\hfill
\subfloat[\scriptsize{Embed weight $\beta$}]{
		\label{Fig:bStudy}%[0.24\textwidth] % width of caption
    \includegraphics[trim = 2mm 0mm 11mm 0mm, clip, width=0.23\textwidth]{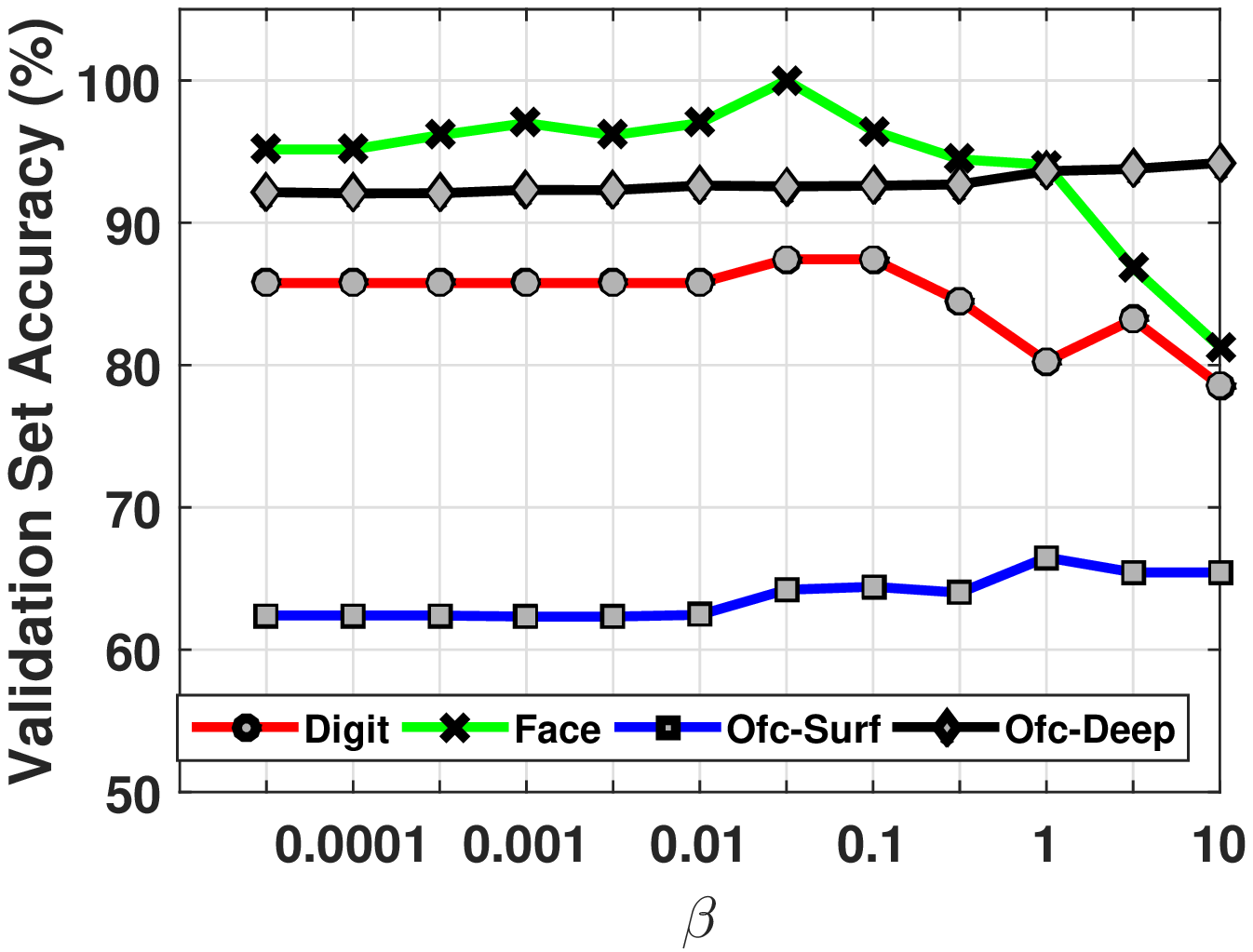}
}%
\hfill
\subfloat[\scriptsize{Regularization $\gamma$}]{
		\label{Fig:gStudy}%[0.24\textwidth] % width of caption
    \includegraphics[trim = 2mm 0mm 11mm 0mm, clip, width=0.23\textwidth]{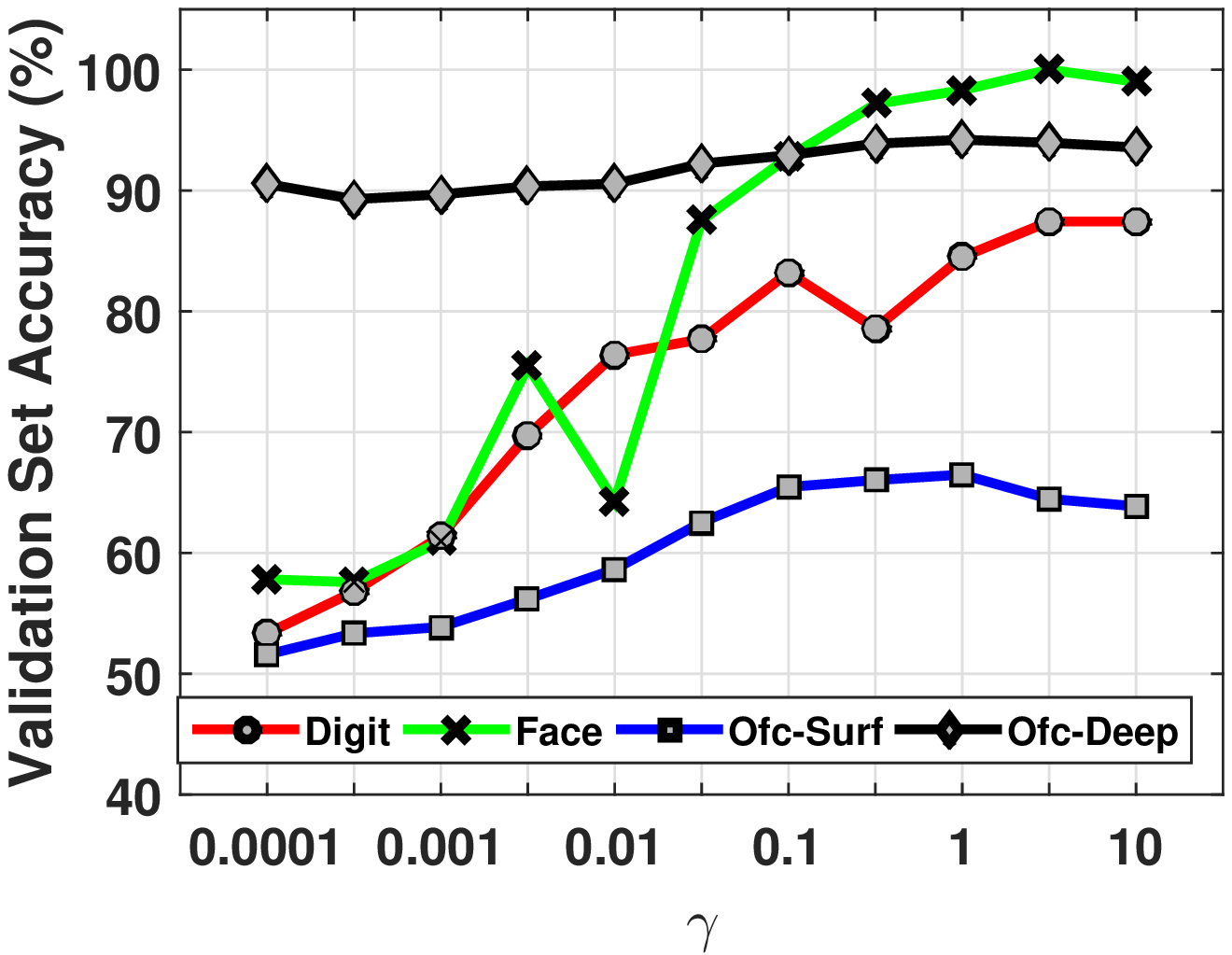}
}%
\caption{\scriptsize{Each figure depicts the accuracies over the validation set for a range of values. When studying a parameter (say $k$), the remaining parameters $(\alpha, \beta, \gamma)$ are fixed at the optimum value.}}
\label{Fig:Exp}
\end{figure}

\section{Experiments}
We compare the NET algorithm with the following baseline and state-of-the-art methods. 
NA (No Adaptation - classifier trained on the source and tested on the target), 
SA (Subspace Alignment \cite{fernando2013unsupervised}), 
CA (Correlation Alignment (CORAL) \cite{sun2015return}),
GFK (Geodesic Flow Kernel \cite{gong2012geodesic}),  
TCA (Transfer Component Analysis \cite{pan2011domain}),  
JDA (Joint Distribution Adaptation \cite{long2013transfer}). 
$\text{NET}_v$ is a special case of the NET algorithm where parameters $(\alpha, \beta, \gamma, k)$, have been estimated using (\ref{Eq:KMM}) (see Sec. \ref{Sec:ModSel}). 
For $\text{NET}^*$, the optimum values for $(\alpha, \beta, \gamma, k)$ are estimated using the target data for cross validation.
%TCA and JDA are the most closely related methods to NET. 
%Transfer Joint Matching (TJM) \cite{long2014transfer}, can be viewed as a specific case of the JDA. 
%While TCA, JDA and TJM use MMD to align the source and target datasets, the NET uses nonlinear embedding and MMD to perform DA.
\begin{table}[t]
\centering
\caption{\scriptsize{Recognition accuracies (\%) for DA experiments on the digit and face datasets. \{\texttt{MNIST}(M), \texttt{USPS}(U), \texttt{CKPlus}(CK), \texttt{MMI}(MM). M$\rightarrow$U implies M is source domain and U is target domain. The best and second best results are in \textbf{bold} and \textit{italic}.}}
\label{Tab:DigitFace}
\resizebox{\linewidth}{!}{%
\begin{tabular}{|c|c|c|c|c|c|c|c|c||c|c|c|c|c|c|c|c|c|}
\hline
\textbf{Expt.} & \textbf{NA} & \textbf{SA} & \textbf{CA} & \textbf{GFK} & \textbf{TCA} & \textbf{JDA} & $\textbf{NET}_v$ & $\textbf{NET}^*$ & \textbf{Expt.} & \textbf{NA} & \textbf{SA} & \textbf{CA} & \textbf{GFK} & \textbf{TCA} & \textbf{JDA} & $\textbf{NET}_v$ & $\textbf{NET}^*$\\ \hline\hline
M$\rightarrow$ U & 65.94 & 67.39 & 59.33 & 66.06 & 60.17 & 67.28 & \textit{72.72} & \textbf{75.39} & CK$\rightarrow$ MM & 29.90 & 31.12 & 31.89 & 28.75 & 32.72 & 29.78 & \textbf{30.54} & \textit{29.97} \\\hline
U$\rightarrow$ M & 44.70 & 51.85 & 50.80 & 47.40 & 39.85 & \textit{59.65} & 61.35 & \textbf{62.60} & MM$\rightarrow$ CK & 41.48 & 39.75 & 37.74 & 37.94 & 31.33 & 28.39 & \textit{40.08} & \textbf{45.83} \\\hline
\textbf{Avg.}	  & 55.32 & 59.62 & 55.07 & 56.73 & 50.01 & 63.46 & \textit{67.04} & \textbf{68.99} & \textbf{Avg.}	  & 35.69 & 35.43 & 34.81 & 33.35 & 32.02 & 29.08 & \textit{35.31} & \textbf{37.90} \\\hline

\end{tabular}
}
\end{table} 

\subsection{Datasets}
\noindent\textbf{\textit{Office-Caltech} datasets}: 
%This is currently the most popular object recognition dataset for DA in the computer vision community. 
This object recognition dataset \cite{gong2012geodesic}, consists of images of everyday objects categorized into 4 domains; \texttt{Amazon}, \texttt{Caltech}, \texttt{Dslr} and \texttt{Webcam}. 
%The \texttt{Amazon} domain has images downloaded from the www.amazon.com website. 
%The \texttt{Dslr} and \texttt{Webcam} domains have images captured using a DSLR camera and a webcam respectively. 
%The \texttt{Caltech} domain is a subset of the Caltech-256 dataset.  
It has 10 categories of objects and a total of 2533 images. 
We experiment with two kinds of features (i) SURF features obtained from \cite{gong2012geodesic}, (ii) Deep features. 
To extract deep features, we use an `off-the-shelf' deep convolutional neural network (VGG-F model \cite{Chatfield14}). 
% The VGG-F network is similar in architecture to the popular AlexNet \cite{krizhevsky2012imagenet}. 
We use the 4096-dimensional features from the $fc8$ layer and apply PCA to reduce the feature dimension to 500. 

\noindent\textbf{\textit{MNIST-USPS} datasets}: We use a subset of the popular handwritten digit (0-9) recognition datasets (2000 images from \texttt{MNIST} and 1800 images from \texttt{USPS} based on \cite{long2014transfer}). 
The images are resized to $16 \times 16$ pixels and represented as 256-dimensional vectors. 

\noindent\textbf{\textit{CKPlus-MMI} datasets}: The CKPlus \cite{lucey2010extended}, and MMI \cite{pantic2005web}, datasets consist of facial expression videos. 
From these videos, we select the frames with the most-intense expression to create the domains \texttt{CKPlus} and \texttt{MMI}, with around 1500 images each and 6 categories viz., \emph{anger, disgust, fear, happy, sad, surprise}. 
We use a pre-trained deep neural network to extract features (see \textit{Office-Caltech}).

\subsection{Results and Discussion}
%We conduct 28 different DA experiments using the above datasets. 
%Each of these is an unsupervised DA experiment with one source domain (data points and labels) and one target domain (data points only). 
%To estimate the optimal parameters $(\alpha, \beta, \gamma, k)$, we weight the source data points using (\ref{Eq:KMM}) in every experiment and select 30\% of the source data points with the largest weights as the validation set. 
For $k$, we explore optimum values in the set $\{10, 20, \ldots, 100, 200\}$.
For $(\alpha, \beta, \gamma)$, we select from $\{0, 0.0001, 0.0005, 0.001, 0.005, 0.01, 0.05, 0.1, 0.5, 1, 5, 10\}$.  
For the sake of brevity, we evaluate and present one set of parameters $(\alpha, \beta, \gamma, k)$, for all the DA experiments in a dataset. 
For all the experiments, we choose 10 iterations to converge to the predicted test/validation labels when estimating $\bM_c$. 
Figure (\ref{Fig:Exp}), depicts the variation in validation set accuracies for each of the parameters. 
We select the parameter value with the highest validation set accuracy as the optimal value in the $\text{NET}_v$ experiments. 

%\subsection{Results and Discussion}
%The parameters $(\alpha, \beta, \gamma, k)$, are fixed to solve for $\bA$ in (\ref{Eq:GenEigen}) and estimate the projected data $\bZ$. 
For fair comparison with existing methods, we follow the same experimental protocol as in \cite{gong2012geodesic}, \cite{long2014transfer}. 
We train a nearest neighbor (NN) classifier on the projected source data and test on the projected target data.  
Table (\ref{Tab:DigitFace}), captures the results for the digit and face datasets. 
Table (\ref{Tab:Object}), outlines the results for the \textit{Office-Caltech} dataset. 
The accuracies reflect the percentage of correctly classified target data points. 
The accuracies obtained with $\text{NET}_v$, demonstrate that the validation set generated from the source data is a good option for validating model parameters in unsupervised DA.
The parameters for the $\text{NET}^*$ experiment are estimated using the target datset; $(\alpha=1, \beta=1,\gamma=1, k=20)$ for the object recognition datasets, $(\alpha=1, \beta=0.01,\gamma=1, k=20)$ for the digit dataset and $(\alpha=0.01, \beta=0.01,\gamma=1, k=20)$ for the face dataset. 
The accuracies obtained with the NET algorithm are consistently better than existing methods, demonstrating the role of nonlinear embedding along with domain alignment. 

\begin{table}[t]
\centering
\caption{\scriptsize{Recognition accuracies (\%) for DA experiments on the \textit{Office-Caltech} dataset with SURF and Deep features. \{\texttt{Amazon}(A), \texttt{Webcam}(W), \texttt{Dslr}(D), \texttt{Caltech}(C)\}. A$\rightarrow$W implies A is source and W is target. The best and second best results are in \textbf{bold} and \textit{italic}.}}
\label{Tab:Object}
\resizebox{\linewidth}{!}{%
\begin{tabular}{|c|c|c|c|c|c|c|c|c||c|c|c|c|c|c|c|c|}
\hline
\multirow{2}{*}{\textbf{Expt.}} & \multicolumn{8}{c||}{\textbf{SURF Features}} & \multicolumn{8}{c|}{\textbf{Deep Features}}\\
\hhline{~----------------}
& \textbf{NA} & \textbf{SA} & \textbf{CA} & \textbf{GFK} & \textbf{TCA} & \textbf{JDA} & $\textbf{NET}_v$ & $\textbf{NET}^*$ & \textbf{NA} & \textbf{SA} & \textbf{CA} & \textbf{GFK} & \textbf{TCA} & \textbf{JDA} & $\textbf{NET}_v$ & $\textbf{NET}^*$\\ \hline\hline
A$\rightarrow$ C & 34.19 & 38.56 & 33.84 & 39.27 & 39.89 & 39.36 & \textit{43.10} & \textbf{43.54} & \textbf{83.01} & 80.55 & \textit{82.47} & 81.00 & 75.53 & \textbf{83.01} & 82.28 & \textbf{83.01} \\\hline
A$\rightarrow$ D & 35.67 & 37.58 & 36.94 & 34.40 & 33.76 & \textit{39.49} & 36.31 & \textbf{40.76} & 84.08 & 82.17 & 87.90 & 82.80 & 82.17 & \textit{89.81} & 80.89 & \textbf{91.08} \\\hline
A$\rightarrow$ W & 31.19 & 37.29 & 31.19 & \textit{41.70} & 33.90 & 37.97 & 35.25 & \textbf{44.41} & 79.32 & 82.37 & 80.34 & 84.41 & 76.61 & 87.12 & \textit{87.46} & \textbf{90.85} \\\hline
C$\rightarrow$ A & 36.01 & 43.11 & 36.33 & \textit{45.72} & 44.47 & 44.78 & 46.24 & \textbf{46.45} & 90.70 & 88.82 & \textit{91.12} & 90.60 & 89.13 & 90.07 & 90.70 & \textbf{92.48} \\\hline
C$\rightarrow$ D & 38.22 & 43.95 & 38.22 & 43.31 & 36.94 & \textit{45.22} & 36.31 & \textbf{45.86} & 83.44 & 80.89 & 82.80 & 77.07 & 75.80 & 89.17 & \textit{90.45} & \textbf{92.36} \\\hline
C$\rightarrow$ W & 29.15 & 36.27 & 29.49 & 35.59 & 32.88 & \textit{41.69} & 33.56 & \textbf{44.41} & 76.61 & 77.29 & 79.32 & 78.64 & 78.31 & \textit{85.76} & 84.07 & \textbf{90.85} \\\hline
D$\rightarrow$ A & 28.29 & 29.65 & 28.39 & 26.10 & 31.63 & 33.09 & \textit{35.60} & \textbf{39.67} & 88.51 & 84.33 & 86.63 & 88.40 & 88.19 & 91.22 & \textit{91.43} & \textbf{91.54} \\\hline
D$\rightarrow$ C & 29.56 & 31.88 & 29.56 & 30.45 & 30.99 & 31.52 & \textit{34.11} & \textbf{35.71} & 77.53 & 76.26 & 75.98 & 78.63 & 74.43 & 80.09 & \textbf{83.38} & \textit{82.10} \\\hline
D$\rightarrow$ W & 83.73 & 87.80 & 83.39 & 79.66 & 85.42 & \textit{89.49} & \textbf{90.51} & 87.80 & \textit{99.32} & 98.98 & \textit{99.32} & 98.31 & 97.97 & 98.98 & \textbf{99.66} & \textbf{99.66} \\\hline
W$\rightarrow$ A & 31.63 & 32.36 & 31.42 & 27.77 & 29.44 & 32.78 & \textit{39.46} & \textbf{41.65} & 82.34 & 84.01 & 82.76 & 88.61 & 86.21 & 91.43 & \textit{91.95} & \textbf{92.58} \\\hline
W$\rightarrow$ C & 28.76 & 29.92 & 28.76 & 28.41 & 32.15 & 31.17 & \textit{32.77} & \textbf{35.89} & 76.53 & 78.90 & 74.98 & 76.80 & 76.71 & \textbf{82.74} & 82.28 & \textit{82.56} \\\hline
W$\rightarrow$ D & 84.71 & 90.45 & 85.35 & 82.17 & 85.35 & \textit{89.17} & \textbf{91.72} & 89.81 & \textit{99.36} & \textbf{100.00} & \textbf{100.00} & \textbf{100.00} & \textbf{100.00} & \textbf{100.00} & \textbf{100.00} & \textit{99.36} \\\hline
\textbf{Avg.}	  & 40.93 & 44.90 & 41.07 & 42.88 & 43.07 & \textit{46.31} & 46.24 & \textbf{49.66} & 85.06 & 84.55 & 85.30 & 85.44 & 83.42 & \textit{89.12} & 88.71 & \textbf{90.70} \\\hline
\end{tabular}
}
\end{table}

\section{Conclusions and Acknowledgments}
We have proposed the NET algorithm for unsupervised DA along with a procedure for generating a validation set for model selection using the source data.
Both the validation procedure and NET have better recognition accuracies than competitive visual DA methods across multiple vision based datasets.
This material is based upon work supported by the National Science Foundation (NSF) under Grant No:1116360. 
Any opinions, findings, and conclusions or recommendations expressed in this material are those of the authors and do not necessarily reflect the views of the NSF.
%\clearpage

%\bibliographystyle{splncs03}
%\bibliography{0007bib}

\end{document}